\documentclass[journal]{IEEEtran}
\usepackage{booktabs}
\usepackage{url}
\usepackage{epsfig}
\usepackage{subfigure}
\usepackage{graphics}
\usepackage{graphicx}
\usepackage{formular}
\usepackage{algorithm}
\usepackage{algorithmic}
\usepackage{multirow}
\usepackage{multicol}
\usepackage{xcolor}
\usepackage{clrscode}
\usepackage{makecell}
\usepackage{CJK}
\usepackage{amsthm}
\usepackage{balance}
\usepackage{times}
\usepackage{amsmath}
\usepackage{amsfonts}
\usepackage{stfloats}
\usepackage{array}
\usepackage{float}
\usepackage{url}
\usepackage{epstopdf}
\usepackage{threeparttable} 
\usepackage[table]{xcolor}
\newtheorem{ruledef}{Rule}

\begin{document}
\title{MediAD: Cross-modal Causal Intervention with Mediator for Alzheimer’s Disease Diagnosis}

\author{
\IEEEauthorblockN{Yutao Jin} \\
\IEEEauthorblockA{Southwest Jiaotong University,  Chengdu, China Email: jinyutao@my.swjtu.edu.cn.} \\
\and 
\IEEEauthorblockN{Haowen Xiao} \\
\IEEEauthorblockA{Southwest Jiaotong University,  Chengdu, China Email: xhaowen@my.swjtu.edu.cn.} \\
\and 
\IEEEauthorblockN{Junyong Zhai} \\
\IEEEauthorblockA{Southeast University,  Nanjing, China Email: jyzhai@seu.edu.cn.} \\
\and 
\IEEEauthorblockN{Yuxiao Li} \\
\IEEEauthorblockA{Sichuan University,  Chengdu, China, Email: wchyuxiaoli@outlook.com.} \\
\and 
\IEEEauthorblockN{Fengmao Lv and Jielei Chu}\\
\IEEEauthorblockA{Southwest Jiaotong University,  Chengdu, China Email: \{fengmaolv, jieleichu\}@swjtu.edu.cn.}
} 

\maketitle
\begin{abstract}
Mild Cognitive Impairment (MCI) serves as a prodromal stage of Alzheimer's Disease (AD), where early identification and intervention can effectively slow the progression to dementia. However, diagnosing AD remains a significant challenge in neurology due to the confounders caused mainly by the selection bias of multi-modal data and the complex relationships between variables. To address these issues, we propose a novel visual-language causality-inspired framework named Cross-modal Causal Intervention with Mediator for Alzheimer’s Disease Diagnosis (MediAD) for diagnostic assistance. Our MediAD employs Large Language Models (LLMs) to summarize clinical data under strict templates, therefore enriching textual inputs. The MediAD model utilizes Magnetic Resonance Imaging (MRI), clinical data, and textual data enriched by LLMs to classify participants into Cognitively Normal (CN), MCI, and AD categories. Because of the presence of confounders, such as cerebral vascular lesions and age-related biomarkers, non-causal models are likely to capture spurious input-output correlations, generating less reliable results. Our framework implicitly mitigates the effect of both observable and unobservable confounders through a unified causal intervention method. Experimental results demonstrate the outstanding performance of our method in distinguishing CN/MCI/AD cases, outperforming other methods in most evaluation metrics. The study showcases the potential of integrating causal reasoning with multi-modal learning for neurological disease diagnosis.
\end{abstract}
\IEEEpeerreviewmaketitle
\section{Introduction}
Alzheimer's Disease (AD) is an irreversible neurodegenerative disorder characterized by progressive cognitive decline and distinct neuropathological alterations in brain tissue. Its clinical manifestations follow a continuous progression from Mild Cognitive Impairment (MCI) to dementia stages: early-stage symptoms predominantly involve episodic memory decline, intermediate-stage symptoms include language dysfunction and executive ability impairment, and late-stage symptoms result in the loss of basic self-care capacity. Studies indicate that initiating treatment during the MCI phase can delay disease progression by 40\%. Therefore, early diagnosis of AD has become a major focus of clinical research.
\begin{figure}
\centering
\includegraphics[width=0.8\columnwidth]{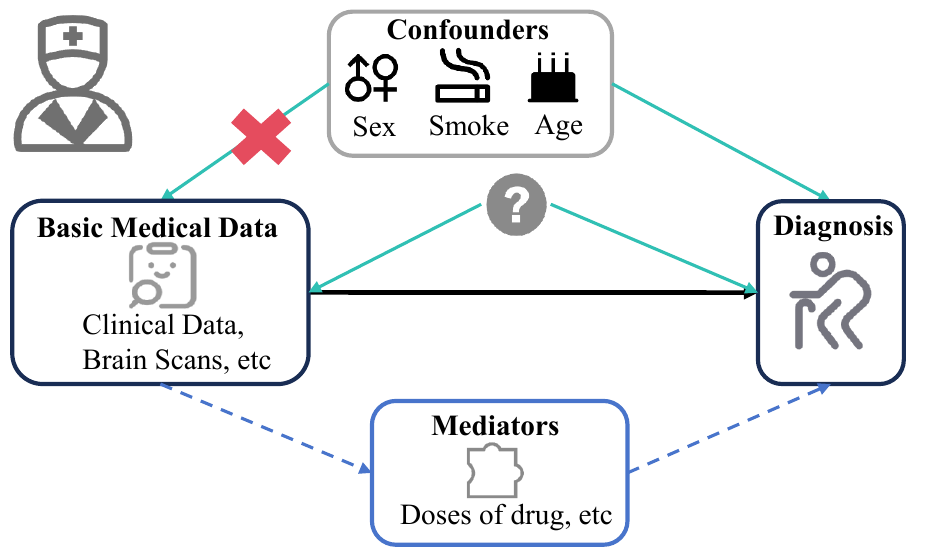}
\caption{Doctors use multi-modal medical data to achieve more accurate diagnoses through mediation analysis and confounding factor screening. Confounding factor screening is often used in scenarios where confounders are observable, whereas mediation analysis is frequently employed for accurate clinical diagnosis in settings where confounders are unobservable.}
\label{fig:intro}
\end{figure}
As shown in Fig.~\ref{fig:intro}, clinical diagnostic methods can be summarized as a chained reasoning process based on medical test data and driven by medical knowledge. Specifically, the diagnostic process of AD usually relies on neuropsychological scales and brain imaging to provide support for basic medical data, and further screening and gradual elimination of confounders~\cite{malec2023causal}. Some works~\cite{yu2023uncovering,andrews2020examining,lee2019mediation,lee2021guideline} have combined with the use of mediation analysis to pursue an accurate clinical diagnosis. Within this process, the exclusion of confounders is a central step in clinical diagnosis. Without intervention, confounders distort the true causal effect, leading to biased estimates that deviate from the true causal relationship. In AD diagnosis, confounders can obscure genuine pathological signals, misrepresenting the mechanisms of the disease. 
Confounders manifest in diverse forms. Creatinine and BMI can simultaneously influence the risk of AD and plasma biomarkers~\cite{pichet2023confounding}, while plasma biomarkers (especially NfL and GFAP) themselves serve as key diagnostic criteria. When assessing certain biomarkers, age can also be a confounder in AD diagnosis~\cite{oh2016dynamic}, as it affects the accumulation of amyloid-$\beta$ in the brain. Amyloid-$\beta$ is a hallmark pathological feature of AD. Demographic imbalances in gender, race, or educational attainment within datasets can lead to delayed clinical manifestation of cognitive decline in populations with higher education levels. Additionally, vascular lesions in AD imaging may be mistaken for specific features of AD~\cite{holland2008spatial}, resulting in elevated false positive rates. The factors mentioned above pose a great challenge for early-stage clinical screening of AD and greatly affect the outcome of AD treatment.

With the recent progress in multi-modal deep learning~\cite{jiaoda3} and LLMs~\cite{jiaoda2}, many studies are attempting to solve the problem of early AD diagnosis by employing these advanced techniques.
Due to the subtle structural changes in the brain of early-stage patients, traditional methods rely on manual pre-selection of lesion regions and fail to integrate multi-modal clinical data (e.g., cognitive scores, genetic information), resulting in limited diagnostic accuracy. 3D-CNN~\cite{3dcnn} proposes to automatically learn lesion regions through hierarchical 3D fully convolutional networks and incorporate multi-modal feature fusion to improve the sensitivity and specificity of early diagnosis.                    
To address the challenge of predicting MCI-to-AD progression, which is often hindered by the scarcity of specific MCI subtype labels (pMCI/sMCI), HOPE~\cite{HOPE} leverages widely available coarse-grained labels (NC/MCI/AD). By employing ordered learning, the model captures the inherent patterns of disease progression, thereby reducing its reliance on subtype-specific annotations.
In addition, due to the lack of interpretability of traditional deep learning models~\cite{chu}, it is difficult to gain clinical trust and medical guidelines and experts' experiences are not effectively incorporated, while purely symbolic systems are difficult to process visual data. NeuroSymAD~\cite{NeuroSymAD} proposes a neurosymbolic framework that emulates the multi-modal diagnostic process employed by medical professionals. The framework employs deep learning to process MRI images and integrates this with symbolic inference to incorporate clinical rules. Furthermore, it utilizes LLMs to automatically generate and refine medical rules, thereby enhancing diagnostic accuracy and interpretability.

However, whether using unimodal approaches~\cite{3dcnn} or multi-modal integration~\cite{mmad} with clinical data, these methods essentially stem from data-driven learning without modeling the causal structure in AD diagnosis. Consequently, such non-causal models are vulnerable to interference from confounders. To address this, we use a Structural Causal Model (SCM)~\cite{scm} to guide AD diagnostic reasoning for multi-modal data sources. Specifically, SCMs implement causal interventions by modeling the causal relationship between input data and diagnostic outcomes, which can effectively mitigate the effects of irrelevant confounders. 

Therefore, we propose the Cross-modal Causal Intervention with Mediator for Alzheimer’s Disease Diagnosis (MediAD) for AD diagnostics, a novel causality-inspired framework that integrates structural causal models to explicitly model the causal relationships between visual/textual features, confounders, and diagnostic outcomes. Prior to training, we utilize Large Language Models (LLMs) to generate structured summaries via designed prompt templates. These summaries, in conjunction with pre-processed medical data, form the textual input for our model. The MRI images, which serve as the visual input, undergo a standardized pre-processing pipeline. They are then processed by a feature extractor, which consists of a 3D Convolutional Neural Network (CNN) and a visual encoder, to extract visual features. Summaries and medical data are fed into different embedding layers, subsequently concatenated, and then passed to a text encoder to yield textual features. Following this, the extracted visual and textual features are jointly fed into a cross-modal Causal Fusion (CF) module to derive an appropriate mediator. Meanwhile, visual and textual features are concatenated to form  multi-modal features. Finally, these multi-modal features are fed into a multi-modal encoder, a Front-Door Adjustment (FDA) module, and a classifier to produce the final diagnostic result.

In general, the main contributions of this paper are as follows:
\begin{itemize}
\item[$\bullet$] We implement LLMs-guided clinical data summarization to enrich textual feature inputs. Utilizing specific prompts, Large Language Models (LLMs) transforms unstructured clinical data into structured summarization. This approach aligns with knowledge distillation techniques, where LLMs act as "teacher models" to generate high-fidelity representations of clinical histories, medication records and behavioral symptoms, etc.
\item[$\bullet$] We propose a causality-inspired AD diagnostic framework based on causal intervention to mitigate confounders' impact on diagnosis, achieving outstanding performance. 
\item[$\bullet$] We introduce a novel mediator generation module leveraging cross-modal fusion. Furthermore, we apply consistency regularization to mediators to enable mediators to better satisfy the front-door criterion, thereby achieving unified front-door adjustment on multi-modal features. This approach improves the accuracy of AD diagnosis.
\item[$\bullet$]We design a Front-Door Adjustment module based on Monte Carlo sampling, which demonstrates promising performance and enhances computational efficiency within the deep learning framework proposed in this paper.
\end{itemize}
The rest of this paper is organized as follows: In Section~\ref{Related}, we review the related research and discuss the commonalities and differences between the proposed MediAD and existing methods. Section~\ref{Method} presents the detailed methodology and theoretical background of our MediAD. Section~\ref{Experiment} presents the experimental results and the visualization results. Finally, Section~\ref{Conclusions} concludes this paper, summarizing MediAD and discussing future work. 
%

\section{Related Work}
\label{Related}
\subsection{Causal Inference}
In recent years, the integration of causal inference and deep learning has become a prominent area of research~\cite{li2024survey}, with notable applications in fields like computer vision~\cite{wang2020visual,wang2024vision}, time series analysis~\cite{Causal-TSF,Causalformer}, disentangled representation learning~\cite{wang2023infodiffusion,Causalvae}, and recommender systems~\cite{llmcausal,causalrec}.

In the theory of causal inference, the presence of confounders often leads to spurious correlations between inputs and outputs~\cite{deshpande2022multi,ma2023multi}, resulting in abnormal model predictions. However, as confounders are inherently unobservable or statistically intractable, causal intervention methods, including backdoor adjustment, front-door adjustment, and counterfactual reasoning, which have become critical tools to mitigate these biases. These approaches demonstrate transformative potential across deep learning domains. 
Causal-TSF~\cite{Causal-TSF} uses causal intervention to mitigate the influence of time-varying confounders. CRA~\cite{VQG} achieves multi-modal deconfounding by employing a back-door intervention on the language and a front-door intervention on the video.~\cite{vqacr} proposed a novel CMCIR framework to mitigate spurious correlations via causal intervention and achieved remarkable performance in event-level visual question answering task.~\cite{mrcrossmodal} sequentially applied front-door adjustments to visual representations and  linguistic representations, significantly eliminating visual and linguistic confounders. Another breakthrough comes from Causal-Aware Attention Mechanism (CaaM)~\cite{Caam}, which integrates causal intervention into attention mechanisms by reweighting attention scores using latent mediator variables, achieving outstanding performance on out-of-distribution (OOD) data compared to standard self-attention. 

These advancements demonstrate the potential of combining causal theory with deep learning and show great research value in improving the performance of deep learning in various domain applications through causal interventions such as front-door adjustment, back-door adjustment, and counterfactual reasoning.
\subsection{Large Language Models}
Large Language Models (LLMs)~\cite{gpt,deepseekvl} are increasingly utilized in medical applications~\cite{medllm,biobert,kgllm}, driven by their remarkable capabilities in semantic understanding and processing. MedBlip~\cite{chen2024medblip} leverages a pre-trained image encoder and a frozen Large Language Models to perform multi-modal AD diagnosis.~\cite{aclllm} introduces an efficient pipeline to generate high-quality clinical notes from patient-doctor dialogues, which combines continued pre-training, supervised fine-tuning (SFT), and a reinforcement learning method named DistillDirect. CLINGEN~\cite{kgllmtextgen} leverages external Knowledge Graphs (KGs) and LLMs to generate high-quality synthetic clinical text in few-shot scenarios.

Drawing inspiration from the success of these studies in utilizing LLMs for assistive clinical diagnosis, our method also employs the LLMs to generate structured summaries under a designed prompt template from clinical data, thereby enriching the model's textual inputs.
\subsection{AD Diagnosis With Deep Learning}
With the advancement of deep learning, its application in assisting the diagnosis of AD and early-stage mild cognitive impairment (MCI) has proliferated, yielding promising results.~\cite{HOPE} proposed a hybrid-granularity ordinal prototype learning (HOPE) method to predict MCI progression by integrating multi-granularity biomarkers and sequential cognitive features.~\cite{nature} demonstrated robust performance in classifying dementia etiologies using extensive multi-modal brain imaging data, leveraging spatial-temporal fusion techniques to capture dynamic functional connectivity patterns in structural MRI (sMRI). Notably, NeuroSymAD is a neurosymbolic framework that combines neural networks with symbolic logic reasoning to enhance both diagnostic accuracy and explainability.~\cite{chebconv} introduces a novel GNN method based on clinical data, achieving promising performance in multiple AD classification tasks.

In the clinical diagnosis of AD, it is common to employ causal inference for a reliable diagnosis. However, these methods fail to account for the confounding effects between the clinical data input and the diagnostic result, as they operate within non-causal frameworks. In contrast, our approach establishes a formal causal framework between the clinical data input and the diagnostic output. We specifically focus on the scenario where confounders are unobserved and utilize the front-door adjustment method to mitigate the impact of confounders on the diagnosis. This ensures the robustness and reliability of the final diagnostic results.

\section{Method}
\label{Method}
In this section, we first introduce the fundamental methods of causal intervention and our proposed Structural Causal Model (SCM) based on the AD diagnosis process. After introducing the front-door adjustment principle, we then detail the cross-modal Causal Fusion (CF) module for mediator generation and the Front-Door Adjustment (FDA) module for the intervention. Finally, we present the overall framework and of our MediAD.
\begin{figure}
\centering
\includegraphics[width=\columnwidth]{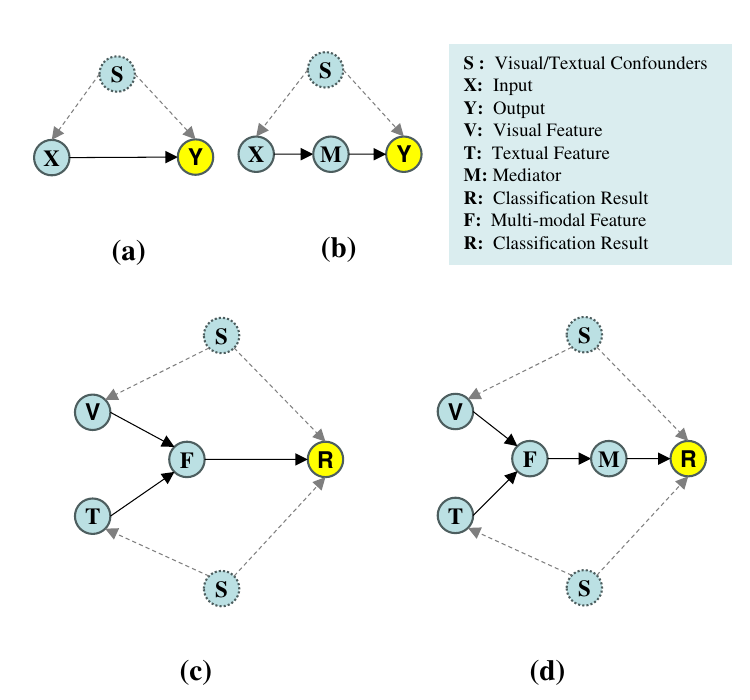}\\
\caption{The structural causal model (SCM). (a) illustrates the causal pathway from input $X$ to output $Y$, where both $X$ and $Y$ are influenced by the confounding factor S. (b) demonstrates the application of front-door adjustment to decompose the causal pathway from $X$ to $Y$ and estimate the causal effect between them when confounders are unobservable. (c) illustrates the structural causal model (SCM) diagram constructed based on our proposed framework. (d) demonstrates the implementation of front-door adjustment in our framework to address unobservable visual and textual confounders by selecting appropriate mediators for multi-modal features. }
\label{fig:SCM}
\end{figure}
\subsection{Causal Intervention}
To formalize the causal intervention mechanism in our methodology, we adopt Judea Pearl's Structural Causal Model (SCM) framework, which is illustrated in Fig.~\ref{fig:SCM}. The SCM provides a mathematical formalism for encoding causal relationships between variables through directed acyclic graphs (DAGs), where nodes represent observable variables and directed edges denote causal dependencies.

In Fig.~\ref{fig:SCM}(a), the causal effects of the input variable $X$ on the output variable $Y$ are formally represented as a directed edge $X \rightarrow Y$ in the structural causal model (SCM).
This causal dependency is mathematically expressed through the conditional probability $P(Y \mid X)$, which quantifies the probabilistic influence of $X$ on $Y$ under observational conditions. However, the presence of biased data introduces the confounders S, which induce spurious correlations between input $X$ and output $Y$. These confounding factors create non-causal dependencies that mislead model predictions, leading to systematic errors in decision-making. Consequently, in the presence of confounders, the true causal effects $X \rightarrow Y$ cannot be reliably calculated through conditional probability alone. For instance, when diagnosing AD based on hippocampal atrophy size observed in brain scans, age may act as a confounder that compromises diagnostic accuracy. 
This occurs because age-related mild hippocampal atrophy in elderly patients, particularly in the absence of supporting biomarkers, could be misattributed to early AD pathological changes. Similarly, when diagnosing AD through neurological assessment in clinical data, age may function as a confounder. Artifacts in brain scan images may act as confounders in AD diagnosis, obscuring the true causal effects between visual features and diagnosis.

An effective method to mitigate confounders is through back-door adjustment. Back-door adjustment employs the $do$-calculus $do(\cdotp)$ to block the backdoor path $Y \leftarrow S \rightarrow X$, and its causal intervention process can be mathematically formalized as follows:
\begin{equation}
\begin{aligned}
P(Y \mid do(X)) = \sum_{s}^{}P(Y \mid X,S=s)P(S=s).
\end{aligned}
\end{equation}
The aforementioned back-door adjustment formula can be interpreted as follows: For each known confounder $S$, the causal effect of $X$ on $Y$ is calculated by computing the weighted average of the effect estimates using the original distribution of $S$ as the weighting factor. 

Since back-door adjustment requires confounders to be observable and measurable, a condition rarely met due to the prevalence of latent confounders. Instead, we employ front-door adjustment to address unobserved confounders, as illustrated in Fig.~\ref{fig:SCM}(b). 
The front-door adjustment introduces a mediator M to eliminate the influence of unmeasurable confounders S, where M blocks the causal pathway $S \rightarrow X \rightarrow M \rightarrow Y$ while satisfying the front-door criterion. The front-door criterion requires the following conditions: (1) M intercepts all directed paths from $X$ to $Y$. (2) There exists no unblocked backdoor path between $X$ and $M$. (3) All backdoor paths from $M$ to $Y$ are blocked by conditioning on X. The causal effect under front-door adjustment can be mathematically formalized as:

\begin{equation}
\label{eq:(2)}
\begin{aligned}
&P(Y \mid do(X = x))\\
&=\sum_{m} P(Y \mid do(M=m)) \, P(M=m \mid X=x), \\
&=\sum_{m} P(M=m \mid X=x)\\
&\times \sum_{x^\prime} P(Y \mid X=x^\prime, M=m) \, P(X=x^\prime).
\end{aligned}
\end{equation}

The aforementioned front-door adjustment formula can be interpreted as follows: $x^\prime$ represents the natural value of variable $X$ under no intervention, while $\sum_{x^\prime}^{}P(Y \mid X=x^\prime, M=m)P(X=x^\prime)$ computes the causal effect of $m$ on $Y$ under the natural distribution of $X$, given $M = m$. Since there is no confounder between $M$ and $X$, the formula calculates the conditional probability $P(m \mid X)$ and weights it by the causal effect of $m$ on $Y$, yielding the total causal effect of $X$ on $Y$ through weighted averaging.

In the AD diagnosis process, data inputs and diagnostic outcomes are affected by observable or unobservable confounders, which can induce spurious input-output correlations. Experienced clinicians employ causal intervention methods to address these challenges, such as progressively eliminating confounding factors and mediation analysis, thereby obtaining accurate and reliable results.
To emulate this clinical reasoning based on causal inference and apply it to a deep learning approach for AD diagnosis, we constructed an idealized Structural Causal Model (SCM) within our MediAD framework. As illustrated in Fig.~\ref{fig:SCM}(c), the proposed SCM is built upon the visual feature $V$, textual feature $T$, and the diagnostic output $R$. The visual and textual feature jointly constitute the multi-modal feature of the model. 
However, the presence of confounders $S$ induces spurious correlations between inputs $V$, $T$ and outcome $R$. 
To address this challenge, we implement front-door adjustment by introducing a mediator $M$ to block the confounding path $S \rightarrow V \rightarrow F \rightarrow M$ and $S \rightarrow T \rightarrow F \rightarrow M$.
Specifically, we employ the cross-modal Causal Fusion (CF) module to estimate the mediator $M$ and apply consistency regularization to ensure the mediator better satisfies the front-door criterion. Subsequently, a Front-Door Adjustment (FDA) module is used to perform front-door adjustment on the multi-modal features.
\subsection{cross-modal Causal Fusion Module}
\begin{figure}
\centering
\includegraphics[width=0.75\columnwidth, keepaspectratio]{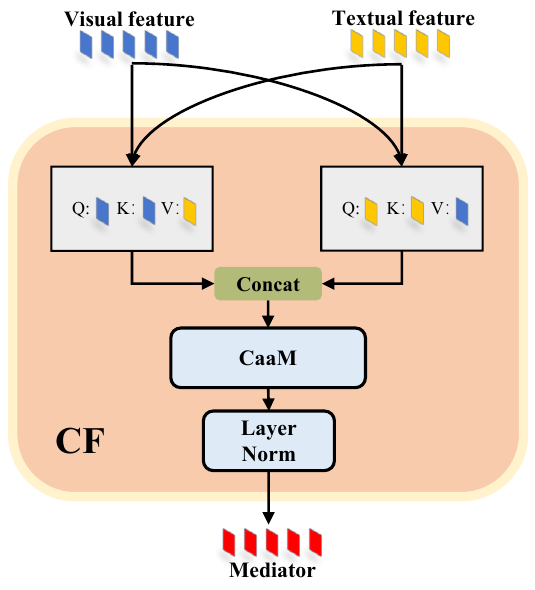}\\
\caption{The structure of cross-modal Causal Fusion (CF) module. Visual feature and textual feature undergo cross-attention computation and are concatenated. The fused features are then passed through the CaaM~\cite{Caam} to further mitigate the effects of confounders, ultimately outputting the refined representation as the mediator.  }
\label{fig:cf}
\end{figure}

 In order to obtain a more reasonable mediator variable $M$, we introduce a cross-modal Causal Fusion (CF) module to generate mediators and perform preliminary removal of confounders, as shown in Fig.~\ref{fig:cf}. The CF module operates through two key computational phases: 
\subsubsection{Cross-modal Attention Computation}
Visual features $f_V$ (Textual features $f_T$) serve as queries and Textual features (Visual features) serve as keys and values in the multi-head attention mechanism respectively for generating features $f_{VTT}$ and $f_{TVV}$ through Multi-Head Attention. $f_{VTT}$ and $f_{TVV}$ are then concatenated to calculate cross-modal features $f_{cm}$. Meanwhile, a small amount of random noise is injected into the visual features (text features), and the same steps are followed to generate $f_{\epsilon}$, simulating the scenario where the input features undergo minor perturbations that do not affect the core causal relationships:
\begin{equation}
\begin{aligned}
 &f_{VTT} = \alpha \cdot MHA(Q_V,K_T,V_T) ,\\
&f_{VTT_{\epsilon}} = \alpha \cdot MHA(Q_V+\epsilon_{V},K_T+\epsilon_{T},V_T+\epsilon_{T}) ,
\end{aligned}
\end{equation}
\begin{equation}
\begin{aligned}
&f_{TVV} = \beta \cdot MHA(Q_T,K_V,V_V) ,\\
&f_{TVV_{\epsilon}} = \beta \cdot MHA(Q_T+\epsilon_{T},K_V+\epsilon_{V},V_V+\epsilon_{V}) ,
\end{aligned}
\end{equation}
\begin{equation}
\begin{aligned}
&f_{cm} =  concat(f_{VTT} , f_{TVV}),\\
&f_{\epsilon} =  concat(f_{VTT_{\epsilon}} , f_{TVV_{\epsilon}}).
\end{aligned}
\end{equation}
Here, $\alpha$ and $\beta$ are trainable parameters of the model. $\epsilon_{V}$ and $\epsilon_{T}$ represent the random noise added to the visual features and textual features. Both $\epsilon_{V}$ and $\epsilon_{T}$ are sampled from a standard normal distribution, i.e., $\epsilon \sim \mathcal{N}(0, 1)$. $MHA$ denotes Multi-Head Attention Layer. 

\subsubsection{Causal Attention Computation}
The cross-modal features obtained in the previous step, as well as the cross-modal features containing noise, are then fed into the causal attention block (CaaM)~\cite{Caam}, an attention module that reduces the influence of confounders in an unsupervised fashion, to further reduce the causal effects of confounders on the mediators:
\begin{equation}
\begin{aligned}
M = LN(CaaM(f_{cm})),
\end{aligned}
\end{equation}
\begin{equation}
\begin{aligned}
M_{\epsilon} = LN(CaaM(f_{\epsilon})),
\end{aligned}
\end{equation}
where $M$ denotes the mediator, $M_{\epsilon}$ denotes the mediator containing noise, and $LN$ is the Layer Normalization~\cite{LN}. The $M$ generated here is not a single latent factor, but rather a set of cross-modal causal clues. The objective is to utilize a neural network to fit an appropriate mediator that satisfies the front-door criterion, thereby enabling the unified front-door adjustment of cross-modal data.

\subsection{Front-door Adjustment}
Previous works~\cite{wang2024vision,chen2025cross,vqacr,navigat} frequently approximate front-door adjustment using modules based on attention calculation and linear network fitting. 
Although this approach can approximate the expectation calculation process in causal intervention through weight allocation, it lacks theoretical and empirical evidence that these methods achieve the effect of a true causal intervention. Furthermore, these methods provide limited discussion on the mediator and the required conditions for satisfying the front-door criterion.
We therefore propose the Front-Door Adjustment (FDA) module based on Monte Carlo sampling to efficiently approximate the causal intervention. We also employ a consistency regularization strategy to constrain the learning of the mediator, ensuring that it better satisfies the front-door criterion.
To calculate the true causal effects between multi-modal feature $F$ and classification outcomes $R$, we first calculate $F$ as follows:
\begin{equation}
\begin{aligned}
F = concat(f_V,f_T).
\end{aligned}
\end{equation}
According to the $do$-calculus, given a causal graph $G$ and any disjoint sets of nodes $X, Y, Z,$ and $W$ within $G$, we have the following rules:
\begin{ruledef}[Insertion/Deletion of Observation]
$P(Y=y | do(X=x),Z=z,W=w) = P(Y=y | do(X=x), Z=z)$ if W is conditionally independent of $Y$ given $X$ and Z in the graph where all arrows pointing into $X$ have been removed.
A key consequence is that an observation $W$ that is conditionally independent of $Y$ can be ignored in the do-calculus expression.
\end{ruledef}
\begin{ruledef}[Action/Observation Exchange]
$P(Y=y | do(X=x), Z=z) = P(Y=y | X=x, Z=z)$ if $Z$ blocks all back-door paths from $X$ to $Y$ in the graph where all arrows into $X$ have been removed.
\end{ruledef}
\begin{ruledef}[Deletion of Intervention]
$P(Y=y | do(X=x), Z=z) = P(Y=y | do(X=x))$ if there are no causal paths from $Z$ to $Y$ in the graph where all arrows into $X$ have been removed.
A key consequence is that if an intervention on $X$ has no causal path to $Y$ once another variable $Z$ is also intervened on, the condition on $Z$ can make the intervention on $X$ irrelevant.
\end{ruledef}
Using these rules, We first derive Equation~(\ref{eq:(2)}) with the SCM in Fig.~\ref{fig:SCM}(d) as follows:
\begin{equation}
\begin{aligned}
\label{eq:(9)}
&P(R|do(F = f)) \\
&= \sum_{M} P(R | do(F), M) P(M|do(F)),\\
&= \sum_{M} P(R | do(F), do(M)) P(M|do(F)),
&\quad\text{(Rule2)}\\
&= \sum_{M} P(R | do(F), do(M)) P(M|F),
&\quad\text{(Rule2)}\\
&= \sum_{M} P(R | do(M)) P(M|F),
&\quad\text{(Rule3)}\\
&= \mathbb{E}_{M \mid F=f}[ P(R \mid do(M)) ]. 
\end{aligned}
\end{equation}
We then use the Normalized Weighted Geometric Mean (NWGM)~\cite{xu2015show} method to estimate $P(R\mid do(M = m))$ as formulated below:
\begin{equation}
\begin{aligned}
\label{eq:(10)}
& P(R\mid do(M = m))  \\
&= \sum_{f'}P(R|do(M), f')P(f'|do(M)),\\
&= \sum_{f'}P(R|M, f')P(f'|do(M)),
&\quad\text{(Rule2)}\\
&= \sum_{f'}P(R|M, f')P(f'),
&\quad\text{(Rule3)}\\
&= \mathbb{E}_{f’}[ P(R \mid M, f’) ], \\
&= \mathbb{E}_{f’}[softmax(g(M, f')) ],\\
&  \approx softmax(\mathbb{E}_{f’}[g( M, f')]),
&\quad\text{(NWGM)}\\
\end{aligned}
\end{equation}
where g(·) is the network mapping function used to model conditional probability $P(R \mid m,f')$. $F$ and $R$ denote multi-modal features and the classification result.

Considering the complexity of the representation space of $f'$, it is computationally prohibitive to calculate the mathematical expectation in Equation~(\ref{eq:(10)}). In addition, methods that use attention mechanisms to approximate front-door adjustment cannot clearly reveal how the model performs the causal intervention process. 
We utilize a Front-Door Adjustment module to address these issues. It uses Monte Carlo sampling to efficiently approximate the mathematical expectation in the causal intervention formula and is built on a more rigorous theoretical foundation.
Specifically, we use the current mini-batch input $\{f_j\}_{j=1}^B$ as the sampling set for feature $f'$.
\begin{equation}
\begin{aligned}
&\mathbb{E}_{f'}[g(M, f')] \approx \frac{1}{B}\sum_{j=1}^Bg(M,f_j).
\end{aligned}
\end{equation}

Therefore, the final form of probability $P(R\mid do(M = m))$ is as follows:
\begin{equation}
\begin{aligned}
P(R\mid do(M = m)) \approx softmax( \frac{1}{B}\sum_{j=1}^Bg(M,f_j)).
\end{aligned}
\end{equation}

We approximate the expectation via Monte Carlo sampling. To ensure computational efficiency, we adopt a single-sample approximation, where we draw a single sample $m \sim P(M \mid F=f)$ to estimate the expectation. The mediator $M$ is generated by our CF module, defined by the function $m=C(f)$. Thus, for the i-th input $f_i$ within a batch, the total causal effect is approximated as:
\begin{equation}
\begin{aligned}
\label{eq:(13)}
&P(R\mid do(F=f_i)) \\
&\approx \mathbb{E}_{M|F=f_i}[softmax( \frac{1}{B}\sum_{j=1}^Bg(M,f_j))]_{M=C(f_i)},\\
&\approx softmax( \frac{1}{B}\sum_{j=1}^Bg(C(f_i),f_j)).
\end{aligned}
\end{equation}

\begin{figure}[t]
\centering
\includegraphics[width=\columnwidth]{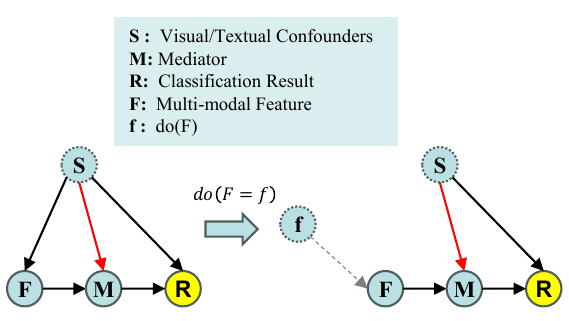}\\
\caption{In the SCM, the expression $do(F=f)$ denotes an intervention on the multi-modal feature $F$, which graphically removes all incoming arrows to $F$. A condition for the front-door adjustment is that there exists no unblocked backdoor path between $F$ and the mediator $M$. The condition will be violated if an unobserved confounder $S$ creates such a backdoor path (e.g., $F \leftarrow S\rightarrow M$). Consequently, the observational probability $P(M|F)$ will not equal the interventional probability $P(M|do(F))$. Conversely, $P(M|do(F)) = P(M|F)$ implies that no such unblocked backdoor path exists.}
\label{fig:pf}
\end{figure}

\begin{figure*}[t]
\centering
\includegraphics[width=\textwidth]{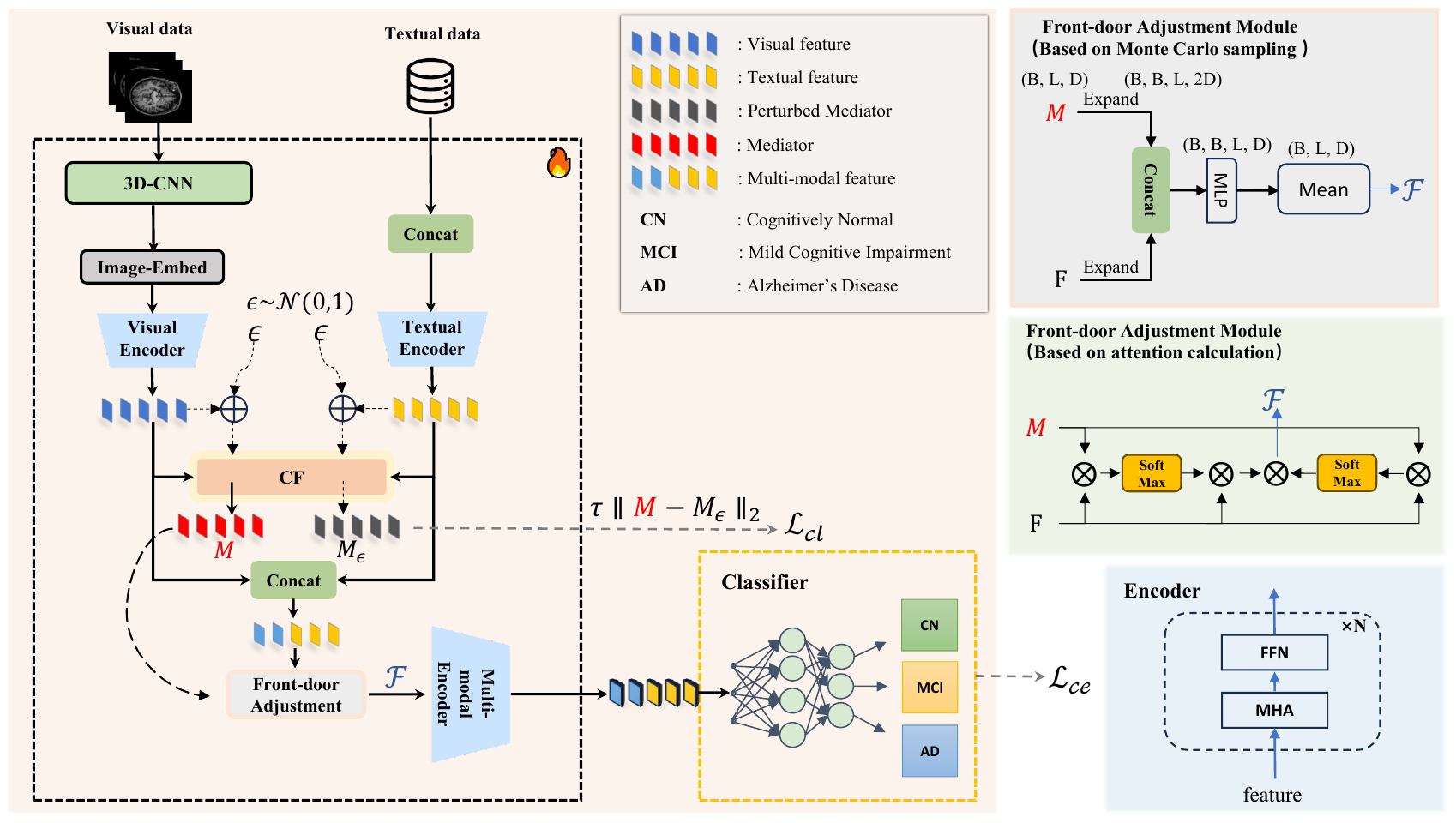}
\caption{An overview of our proposed MediAD. MRI/fMRI scans serve as the visual input, while the textual input consists of summaries generated by Large Language Models (LLMs) and structured clinical data. Visual features are extracted using a 3D-CNN. Within the MediAD framework, these visual and textual features are processed by the cross-modal Causal Fusion (CF) module to generate mediators. The mediators are optimized with a consistency regularization loss ($\mathcal{L}_{cl}$). The multi-modal features and the optimized mediators are then fed into the Front-Door Adjustment (FDA) module to mitigate the effect of confounders. Finally, the refined features from this module are passed to a classifier for final diagnosis.}
\label{fig:overall}
\end{figure*}
\subsection{Textual Data Generation}
\begin{figure*}
\centering
\includegraphics[width=\textwidth]{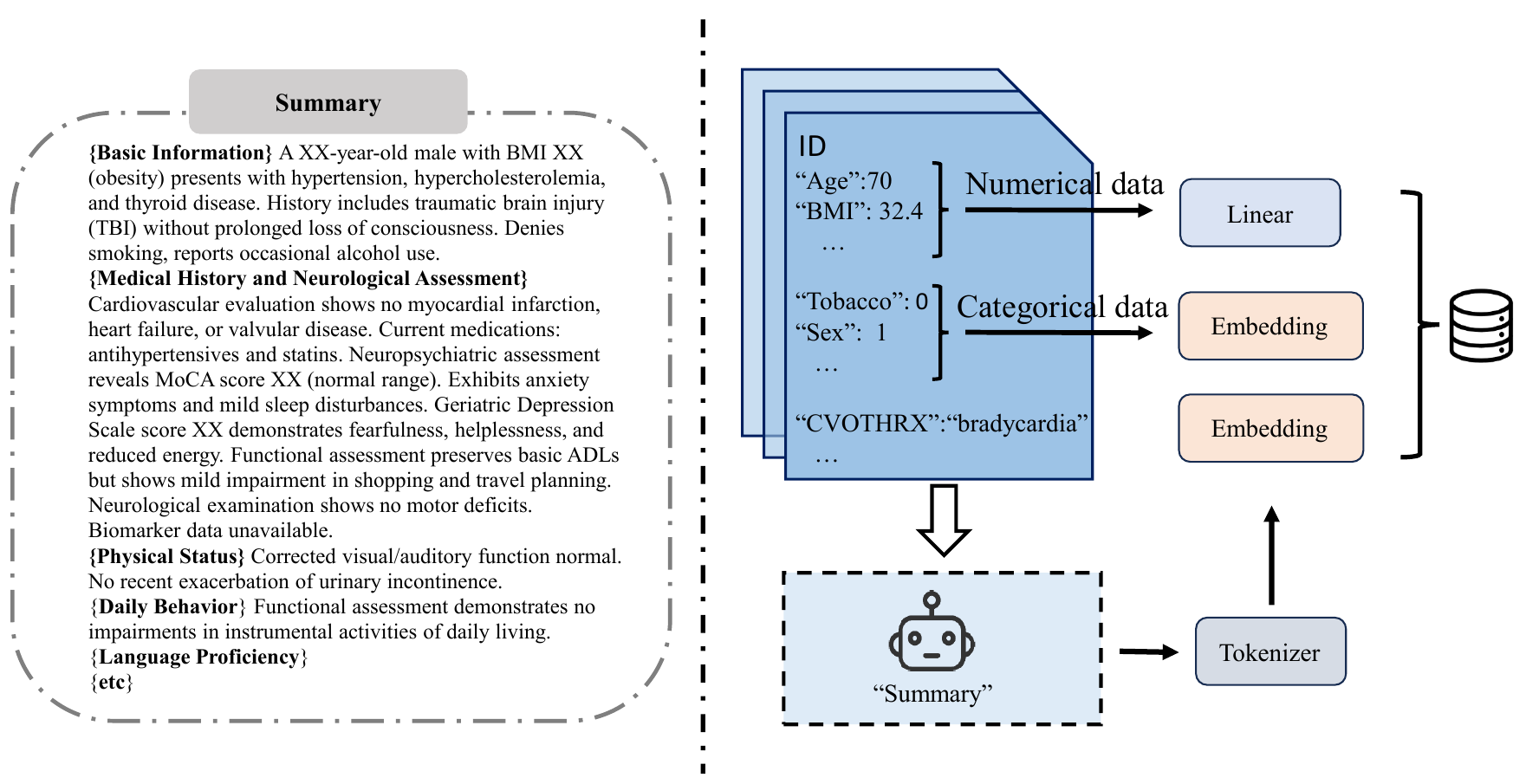}\\
\caption{Textual data processing pipeline. Structured textual data is input into LLMs with designed prompts to produce a clinical summary, which is then tokenized and fed into an embedding layer. Concurrently, the original textual data, pre-classified into numerical and categorical data, is passed through a linear layer and an embedding layer, respectively. The resulting features are then combined with the processed summary to jointly constitute the model's textual input.}
\label{fig:template}
\end{figure*}

Fig.~\ref{fig:template} illustrates our textual data processing pipeline and a general summary generated by LLMs. Each volunteer's clinical data are processed via parallel paths: First, the structured data are partitioned into numerical data and categorical data, which are fed into a Linear layer and an Embedding layer, respectively. To enrich the feature inputs of the model's textual data and integrate unstructured medical information, all data types are input into the LLMs to generate a structured summary. This summary is then tokenized and processed by another Embedding layer. Finally, the features derived from these components (i.e., the processed numerical data, categorical data, and the summary) are jointly concatenated to form the complete textual input for the proposed MediAD.

To achieve this, we employ a structured prompt template and a standardized volunteer information summarization framework, ensuring the comprehensiveness and precision of the LLMs-generated summary.
This template systematically captures critical dimensions of volunteer profiles through the following aspects:
\textbf{Basic Information}, \textbf{Medical History and Neurological Assessment}, \textbf{Physical status}, \textbf{Daily Behavior}, \textbf{language proficiency}, \textbf{etc}.
When encountering unrecorded medical data, we configure LLMs to skip processing these cases.
The specific prompt templates are provided in the Appendix.

\subsection{Framework}
The core framework of MediAD is shown in Fig.~\ref{fig:overall}. Our framework comprises six core components: a visual encoder, a textual encoder, a cross-modal Causal Fusion (CF) module, a Front-Door Adjustment (FDA) module, a multi-modal feature encoder, and a classifier.
We first employ a 3D-CNN~\cite{tran2015learning} as a feature extractor for MRI brain scans. Simultaneously, we employ structured prompt templates to guide DeepSeek~\cite{ds} in generating comprehensive summaries from clinical data.  To further enhance the representation quality of 3D brain scans, we utilize a Vision Transformer (ViT) with image patch embedding. For textual data, we employ a feature extraction framework consisting of a SimpleTokenizer and a standard Transformer~\cite{attn} encoder to extract the textual feature.
The visual and textual features themselves and their versions containing a small amount of random noise are simultaneously fed into the CF module to obtain the mediator $M$ used for front-door adjustment and $M_{\epsilon}$ containing random noise. $M_{\epsilon}$ and $M$ are regularized for consistency using the loss function $\mathcal{L}_{cl}$, which helps the mediator learned by the CF module to better satisfy the front-door criterion.
Subsequently, the multi-modal features and the mediator $M$ are fed into the Front-Door Adjustment (FDA) module to mitigate the influence of confounders. Finally, the resulting purified features are passed into the multi-modal encoder and then to the classifier to generate the final prediction results.

Since the front-door adjustment of input $F$ must satisfy that there is no unblocked backdoor path between $F$ and $M$, i.e., there is no backdoor path such as $F \leftarrow S \rightarrow M$. This is equivalent to $P(M \mid do(F)) = P(M \mid F)$. Although we have preliminarily implemented this condition in our framework through the CF module, in actual training, $M$ may still learn pseudo-correlations related to $S$ in $F$. Therefore, we add $\mathcal{L}_{cl}$ to the loss function to perform consistency regularization on $M$ and $M_{\epsilon}$.
\begin{equation}
\begin{aligned}
    \mathcal{L}_{cl} = \tau \lvert\lvert M - M_{\epsilon} \rvert\rvert_2,  \\
\end{aligned}
\end{equation}
where $\tau = 0.0005$ is a hyper-parameter, $M$ denotes the mediator generated by $f$ and $M_{\epsilon}$ denotes the mediator generated by $f_{\epsilon}$. By constraining the learning process from $F$ to $M$, the learned mediators further satisfy the conditions required for front-door adjustment.
Therefore, the overall loss function of MediAD is as follows:
\begin{equation}
\begin{aligned}
    \mathcal{L} = \mathcal{L}_{ce} + \mathcal{L}_{cl},\\
\end{aligned}
\end{equation}
where $\mathcal{L}_{ce}$ denotes Cross-Entropy loss.

\subsection{Algorithm}
\begin{algorithm}
\caption{Training procedure of MediAD.} 
\label{alg:train}
{\bf Input:} MRI / fMRI data $v \in \mathbb{R}^{h\times w\times d}$, textual data $t$, labels $l$\\
{\bf Output:} Parameters $\theta$ of MediAD

\begin{algorithmic}[1]
    \FOR{batch $(v, t, l)$ in loader}
    \STATE $v' \gets \text{3D-CNN}(v)$
    \STATE $f_v \gets \text{Visual\_Encoder}(v')$
    \STATE $f_t \gets \text{Textual\_Encoder}(t)$ 
    \STATE $F \gets \text{concat}(f_v, f_t)$
    \STATE $M \gets \text{CF}(f_v, f_t)$   
    \STATE $M_{\epsilon} \gets \text{CF}(f_v+\epsilon, f_t+\epsilon)$   
    \STATE $f_{mm} \gets \text{FDA}(F, M)$  \quad //Front-door Adjustment
    \STATE $f_{out} \gets \text{Multi-modal\_Encoder}(f_{mm})$
    \STATE $y_p \gets \text{Classifier}(f_{out})$ 
    \STATE $\mathcal{L}_{cl} \gets \tau \lvert\lvert M - M_{\epsilon} \rvert\rvert_2$
    \STATE $\mathcal{L} \gets \mathcal{L}_{ce} + \mathcal{L}_{cl}$ 
    \STATE update model $\theta$ to minimize $\mathcal{L}$
    \ENDFOR
\end{algorithmic}
\end{algorithm}
Algorithm~\ref{alg:train} presents the training process of the proposed MediAD. Inspired by clinical diagnostic reasoning, MediAD offers a novel multi-modal deep learning framework for AD diagnosis that is formally based on causal intervention.

\begin{table}
\centering
\caption{Basic information of the dataset used in the experiment.}
\label{tab:data}
\begin{tabular}{cccc} 
\toprule
\multirow{2}{*}{\textbf{Dataset}} & \multicolumn{3}{c}{\textbf{MediAD(ours)}} \\
\cline{2-4}
 & \#Group & \#Count & (\%) \\
\midrule
 & NC & 270 & 50.2\%\\
\textbf{ADNI}& MCI & 145 & 26.5\%\\
 & AD & 157 & 23.3\%\\
 & ALL& 572 & 100.0\% \\
\hline
 & NC & 500 & 31.5\%\\
\textbf{NACC}& MCI & 500 & 31.5\%\\
 & AD & 434 & 37.0\%\\
 & ALL& 1434 & 100.0\%\\
\bottomrule
\end{tabular}
\end{table}

\begin{table*}[!htbp]
\renewcommand{\arraystretch}{1.3}
\centering
    \begin{threeparttable}
\caption{Performance of various methods in multi-class classification. The best results are highlighted in bold.}
\label{tab:3class_nacc}
  \begin{tabular}{ccccccccccc}  
    \toprule
    \multirow{2}{*}{\textbf{Method}} & \multicolumn{5}{c}{Internal test on NACC} & \multicolumn{5}{c}{External test on ADNI}\\
    \cline{2-6} \cline{7-11}
     & ACC (\%) & F1 (\%) & Precision (\%) & Recall (\%) & AUC (\%) & ACC (\%) & F1 (\%) & Precision (\%) & Recall (\%) & AUC (\%)\\
    \midrule
    MedBlip                & 83.30 & -     & -     & -     & -              & -     & -     & -     & -       & -             \\
    MediAD$^{\dagger}$     & 85.26 &  85.51&  85.61& 85.42 & \textbf{95.47} & 67.66 & 63.30 & 63.19 & 63.42 & \textbf{79.45}\\
    \rowcolor{gray!20}
    MediAD$^{\ddagger}$  & \textbf{85.69} & \textbf{85.90} & \textbf{86.16} & \textbf{85.63} & 94.83 
                    & \textbf{68.33} & \textbf{68.85} & \textbf{69.38} & \textbf{68.33} & 76.08\\
    \bottomrule
  \end{tabular}
  \begin{tablenotes}
        \footnotesize
        \item $^{\dagger}$ Indicates method that utilizes the Front-door Adjustment using modules based on attention calculations and linear network fitting. $^{\ddagger}$ Indicates method that utilizes the Front-door Adjustment using modules based on Monte Carlo sampling.
      \end{tablenotes}
  \end{threeparttable}
\end{table*}

\begin{table*}[!htbp]
\renewcommand{\arraystretch}{1.3}
\centering
    \begin{threeparttable}
\caption{Performance of various methods in multi-class classification.}
\label{tab:3class_adni}
  \begin{tabular}{ccccccccccc}  
    \toprule
    \multirow{2}{*}{\textbf{Method}} & \multicolumn{5}{c}{Internal test on ADNI} & \multicolumn{5}{c}{External test on NACC}\\
    \cline{2-6} \cline{7-11}
     & ACC (\%) & F1 (\%) & Precision (\%) & Recall (\%) & AUC (\%) & ACC (\%) & F1 (\%) & Precision (\%) & Recall (\%) & AUC (\%)\\
    \midrule
    \texttt{ViT}             & 58.10 & 58.00 & 58.50 & 57.40 & 59.30 & 43.60 & 43.50 & 45.10 & 43.70 & 64.10  \\
    \texttt{OR-CNN}          & 65.70 & 66.90 & 69.00 & 64.90 & 66.90 & 61.70 & 60.80 & 60.70 & 60.80 & 61.40   \\
    \texttt{ADRank}          & 67.00 & 67.90 & 68.40 & 65.70 & 67.90 & 63.40 & 62.10 & 63.00 & 60.90 & 63.00   \\
    \texttt{CORF}            & 67.80 & 68.10 & 69.00 & 67.20 & 68.10 & 64.10 & 62.80 & 63.20 & 62.10 & 63.50   \\
    \texttt{HOPE}            & 72.00 & 71.10 & 71.30 & 71.00 & 71.30 & 67.60 & 66.40 & 66.10 & 67.20 & 66.80  \\
    \texttt{MedBlip}         & 78.70 & - & - & - & -   & - & - & - & - & -\\
    MediAD$^{\dagger}$  & 88.62 & 86.95 & 87.77 & 86.18 & 98.07 & 69.99 & 71.30 & 72.67 & 69.99 & 85.20\\
    \rowcolor{gray!20}
    MediAD$^{\ddagger}$  & \textbf{90.38} & \textbf{90.22} & \textbf{91.48} & \textbf{89.05} & \textbf{97.06} & \textbf{75.00} & \textbf{75.53} & \textbf{76.07} & \textbf{75.00} & \textbf{88.13}\\
    \bottomrule
  \end{tabular}
  \end{threeparttable}
\end{table*}

\begin{table*}[!htb]
\renewcommand{\arraystretch}{1.3}
\setlength{\tabcolsep}{4pt}
\caption{Performance of MediAD in binary-class classification on NACC dataset.}
\label{tab:nacc2class}
\centering
  \begin{tabular}{cccccccccccccccccccc}  
    \toprule
    \multirow{2}{*}{\textbf{Method}} & \multicolumn{5}{c}{\textbf{CN/AD}} & ~ & \multicolumn{5}{c}{\textbf{MCI/AD}} & ~ & \multicolumn{5}{c}{\textbf{CN/MCI}}\\
    \cline{2-6}\cline{8-12}\cline{14-18}
     & ACC & F1 & Precision & Recall & AUC & ~ & ACC & F1 & Precision & Recall & AUC & ~ & ACC & F1 & Precision & Recall & AUC\\
    \midrule
    \rowcolor{gray!20}
    MediAD$^{\ddagger}$ & \textbf{96.03} & \textbf{94.89} & \textbf{95.89} & \textbf{95.90} & \textbf{99.04} 
    &~ & \textbf{89.82} & \textbf{89.97} & \textbf{89.75} & \textbf{89.88} & \textbf{94.29} 
    & ~ & \textbf{85.48} & \textbf{85.84} & \textbf{86.22} & \textbf{86.22} & \textbf{91.94} \\
    \bottomrule
  \end{tabular}
\end{table*}

\begin{table*}
\renewcommand{\arraystretch}{1.3}
\setlength{\tabcolsep}{4pt}
\centering
  \caption{Performance of various methods in binary-class class classification on ADNI dataset.}
  \label{tab:adni2class}
  \begin{tabular}{cccccccccccccccccccc}  
    \toprule
    \multirow{2}{*}{\textbf{Method}} & \multicolumn{5}{c}{\textbf{CN/AD}} & ~ & \multicolumn{5}{c}{\textbf{MCI/AD}} & ~ & \multicolumn{5}{c}{\textbf{CN/MCI}}\\
    \cline{2-6}\cline{8-12}\cline{14-18}
    & ACC & F1 & Precision & Recall & AUC & ~ & ACC & F1 & Precision & Recall & AUC & ~ & ACC & F1 & Precision & Recall & AUC\\
    \midrule
    \texttt{3D-ResNet}& 85.67 & 90.46 & 86.16 & 95.25 & 93.36&~& - & - & - & - & -&~& - & - & - & - & -\\
    \texttt{NeuroSysAD}& 88.58 & 92.15 & 89.97 & 94.44 & 92.56&~& - & - & - & - & -&~& - & - & - & - & -\\
    \texttt{ChebConv}& \textbf{99.00} & \textbf{99.00} & - & - & 99.00&~ & 93.00 &93.00 & - & - & 96.00&~& 78.00 & 76.00& - & - & 83.00\\
    \rowcolor{gray!20}
    MediAD$^{\ddagger}$ & 97.42 & 96.86 & \textbf{96.51} & \textbf{97.24} & \textbf{99.79} 
    &~& \textbf{94.69} & \textbf{93.70} & \textbf{95.15} & \textbf{94.37} & \textbf{97.61} 
    &~& \textbf{89.39} & \textbf{88.91} & \textbf{88.73} & \textbf{89.14} & \textbf{95.84} \\
    \bottomrule
  \end{tabular}
\end{table*}

\section{Experiment}
\label{Experiment}
\subsection{Experimental Setup}

\begin{figure*}
\centering
\includegraphics[width=0.9\textwidth]{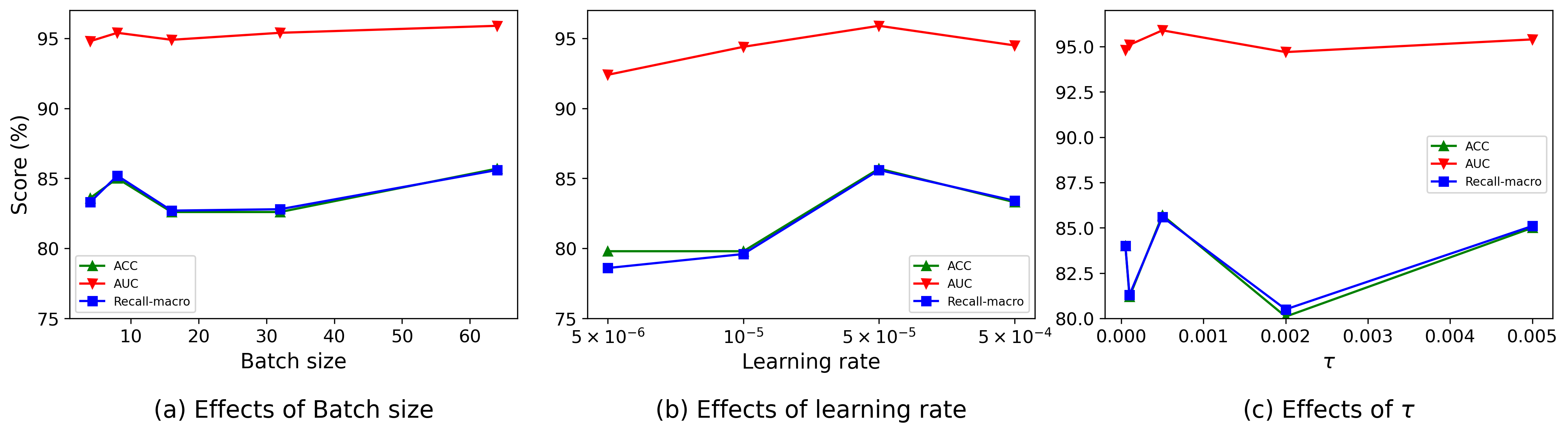}\\
\caption{Sensitivity analysis of the key hyper-parameter on the proposed MediAD.}
\label{fig:sensitive}
\end{figure*}

\subsubsection{Datasets}
Our experiment used two AD research databases: the National Alzheimer's Coordinating Center (NACC)~\cite{nacc} and Alzheimer's Disease Neuroimaging Initiative (ADNI)~\cite{adni} datasets. Both repositories provide multi-modal neuroimaging, clinical, and biomarker data essential for investigating AD progression mechanisms. ADNI synthesizes multi-modal data from 63 research centers globally to elucidate AD pathological mechanisms and identify early diagnostic biomarkers. NACC consolidates multi-center clinical-pathological data across 39 U.S. Alzheimer's Disease Research Centers (ADRCs), providing robust infrastructure for dementia research. 

The experiment utilized 5-fold cross-validation on neuroimaging and clinical data from 2,388 participants in the NACC dataset and participants in the ADNI dataset. In this approach, the data was randomly partitioned into five equally sized folds. The details of our sample selection from the NACC and
ADNI datasets are shown in Table \ref{tab:data}. The model was trained and validated on four folds and tested on the remaining fold, with this process repeated five times to ensure a robust evaluation.
We conducted independent training and testing procedures for the ADNI and NACC datasets to ensure rigorous evaluation of model generalizability. The total number of samples used to train our model with the ADNI dataset is smaller than HOPE and NeuroSysAD.
We present the specific data selection details and cross-validation results in the Appendix. 

\subsubsection{Preprocessing Pipeline}
We selected MRI scans from the NACC dataset and ADNI as visual inputs. All MRI scans acquired from participants underwent preprocessing steps, including spatial resampling, intensity normalization, and automated foreground extraction. All volumetric data were uniformly resized to 128×128×128.
The clinical data was first pre-processed into a structured JSON format and subsequently categorized as categorical, numerical, and textual data. The clinical data was then fed into DeepSeek~\cite{ds} to generate summaries. The template shown in Fig.~\ref{fig:template} was utilized to constrain the structure of the LLMs' output. The categorical data, the numerical data, and  the summaries jointly constitute the textual input of the model.

\subsubsection{Implementation Settings}
The methods in this paper were implemented with the PyTorch library~\cite{paszke2019pytorch}.
The proposed MediAD model was trained on a single NVIDIA H20 GPU platform using an AdamW optimizer~\cite{adamw}. We set the initial learning rate to 5e-5 with cosine annealing scheduling (warmup ratio = 0.1). The weight decay rate of AdamW was set to 0.1 and the batch size was set to 64. Both the visual encoder and textual encoder contain 4 standard transformer~\cite{attn} encoder blocks, while the Multi-modal Encoder contains 2 standard transformer encoder blocks.

\subsubsection{Evaluation Metrics}
We performed the CN/MCI/AD triple classification task on both the ADNI and the NACC datasets, and the CN/AD binary classification task on the NACC dataset.
Five evaluation metrics were used in our task: classification accuracy (ACC), area under the receiver operating characteristic curve (AUC), F1-score, Recall and Precision. 

\subsection{Qualitative Studies}
To validate the effectiveness of the proposed MediAD framework, we demonstrate outstanding performance in most metrics on different datasets in Table~\ref{tab:3class_nacc}, Table~\ref{tab:3class_adni}, Table~\ref{tab:nacc2class}, Table~\ref{tab:adni2class}. For the three-class classification experiments, we utilized the classification results from HOPE~\cite{HOPE} and MedBLIP~\cite{chen2024medblip} as the comparative benchmark and we utilized the classification results from NeuroSysAD~\cite{NeuroSymAD} for the binary classification experiments. 

In detail, our MediAD method achieves better performance compared to previous works~\cite{HOPE,ORCNN,3dreskoro,adrank,CORF,vit}.
Table~\ref{tab:3class_nacc} and Table~\ref{tab:3class_adni} present the performance of our proposed MediAD on the NACC and ADNI datasets, respectively, detailing the results for internal and external evaluations.
Within the NACC internal test cohort, our MediAD$^\ddagger$ (based on Monte Carlo sampling) achieved an ACC 2.39\% higher than MedBlip and 0.43\% higher than MediAD$^\dagger$ (based on attention calculation).
Within the ADNI internal test cohort, MediAD$^\ddagger$ achieves an ACC 14.5\% higher than MedBlip and 0.6\% higher than MediAD$^\dagger$.

Table~\ref{tab:nacc2class} and Table~\ref{tab:adni2class} demonstrate the performance of MediAD$^\ddagger$ on binary classification tasks. The model exhibits higher discriminability between the AD and CN classes, while the distinction between MCI and either AD or CN is relatively lower.

\begin{table}[!htbp]
\renewcommand{\arraystretch}{1.3}
\caption{Ablation performance of MediAD on NACC dataset.}
\centering
\label{tab:ablation}
  \begin{tabular}{cccccc}  
    \toprule
    \multirow{2}{*}{\textbf{Method}} & \multicolumn{5}{c}{\textbf{NACC}} \\
    \cline{2-6}
    & ACC & F1 & Precision & Recall & AUC \\
    \midrule
     \rowcolor{gray!20}
    MediAD$^{\ddagger}$ & \textbf{85.69} & \textbf{85.90} & \textbf{86.16} & \textbf{85.63} & 94.83 \\
        
        \texttt{w/o $\mathcal{L}_{cl}$}          & 85.13 & 85.47 & 85.82 & 85.25 & 95.07\\
        \texttt{w/o $\mathcal{L}_{cl}$$\And$FDA} & 84.64 & 84.94 & 85.13 & 84.74 & \textbf{95.25}\\
        \texttt{w/o LLMs}                         & 81.92 & 82.11 & 82.08 & 82.16 & 92.65\\
    \bottomrule
  \end{tabular}
\end{table}

\subsection{Ablation Studies}
As demonstrated in Table \ref{tab:ablation}, we performed controlled ablation experiments to quantitatively assess the contribution of different components in the proposed MediAD. 
After removing $\mathcal{L}_{cl}$, the model's performance decreased in ACC, F1, Recall and Precision. This decline is attributed to the fact that mediators learned through consistency regularization are better able to help the model fully implement the front-door criterion for stable and reliable inference. Further removal of FDA resulted in a further decline in the model's performance, indicating the critical role of front-door adjustment strategies in AD classification tasks.
Additionally, we removed LLMs-generated summaries from the input textual data to validate the effectiveness of LLMs-generated summaries. Experimental results demonstrate that LLMs-generated summaries effectively enhance model performance.

These results indicate that introducing an appropriate front-door adjustment into our proposed MediAD can effectively mitigate the influence of confounders and improve diagnostic accuracy.

\subsection{Sensitivity Analysis}
We conducted a sensitivity analysis on three key hyperparameters: learning rate, batch size, and $\tau$. Sensitivity experiments were performed on the NACC dataset, which was split into training, validation, and test sets at a 6:2:2 ratio. We evaluated each hyperparameter independently, fixing the remaining ones to their empirically determined optimal values, i.e., a learning rate of 0.00005, a batch size of 64, and $\tau$ of 0.0005. Fig.~\ref{fig:sensitive}(b) shows that all metrics increase with the learning rate before starting to decline, achieving optimal performance when the learning rate is around 0.00005. Fig.~\ref{fig:sensitive}(a) and Fig.~\ref{fig:sensitive}(c) show that the best performance across all metrics is achieved with a batch size of 64 and $\tau$ of 0.0005, respectively.

\subsection{Visualization}
\begin{figure*}[t]
  \centering
  \begin{minipage}[t]{0.22\textwidth}  
    \centering
    \includegraphics[width=\linewidth, keepaspectratio]{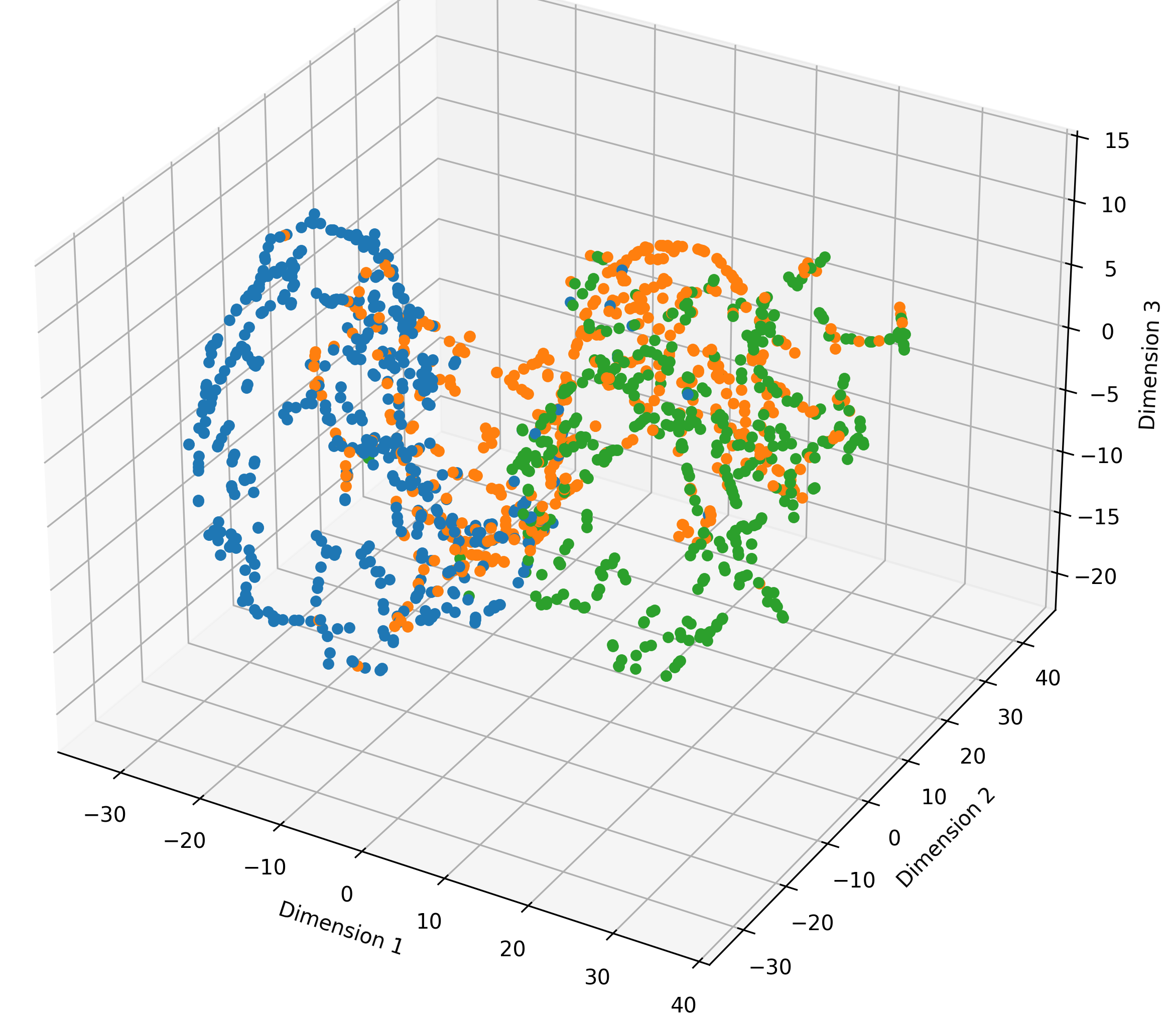}
  \end{minipage}
  \hfill  
  \begin{minipage}[t]{0.22\textwidth}  
    \centering
    \includegraphics[width=\linewidth, keepaspectratio]{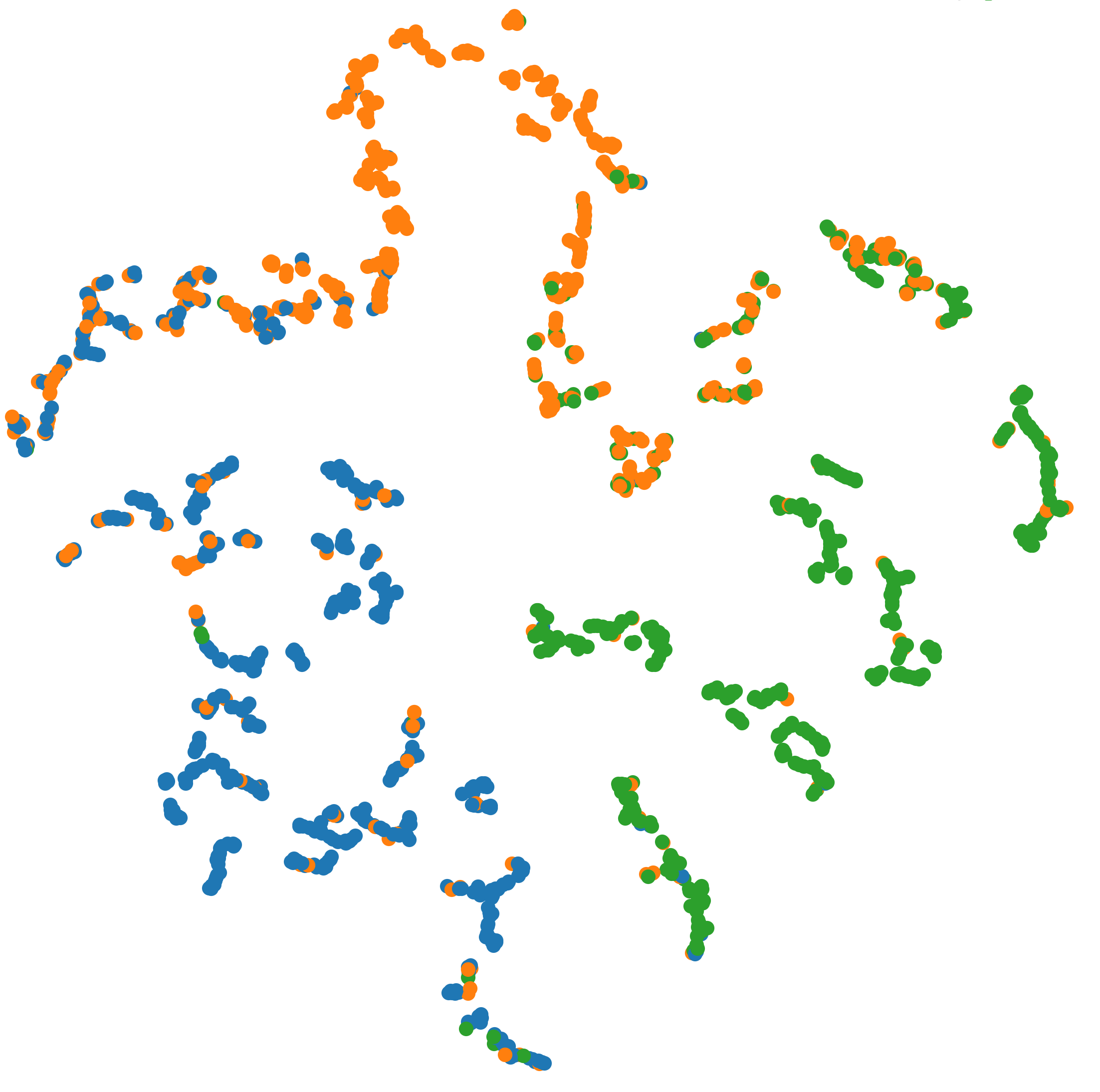}
  \end{minipage}
  \hfill  
  \begin{minipage}[t]{0.22\textwidth}
    \centering
    \includegraphics[width=\linewidth, keepaspectratio]{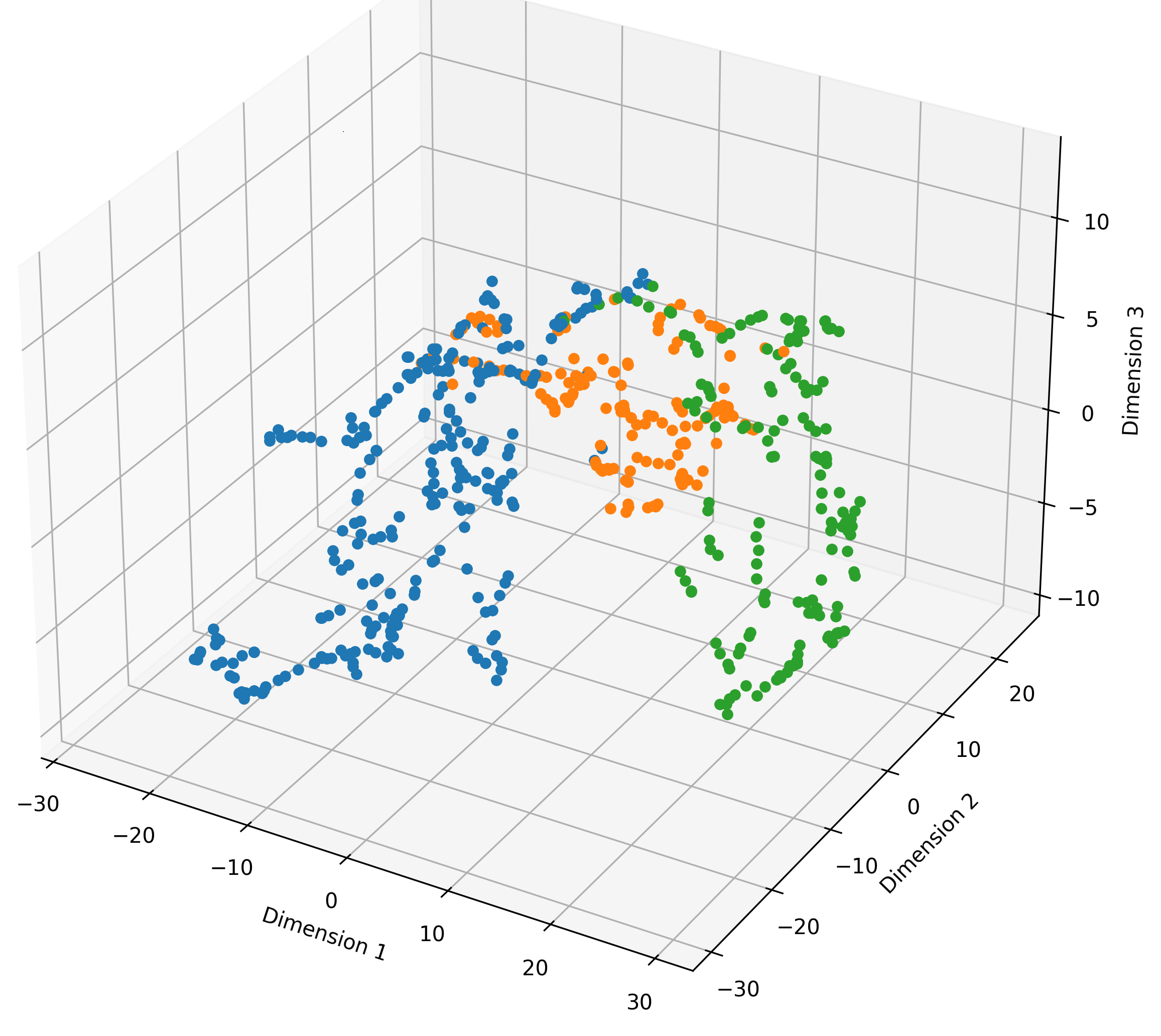}
  \end{minipage}
  \begin{minipage}[t]{0.22\textwidth}
    \centering
    \includegraphics[width=\linewidth, keepaspectratio]{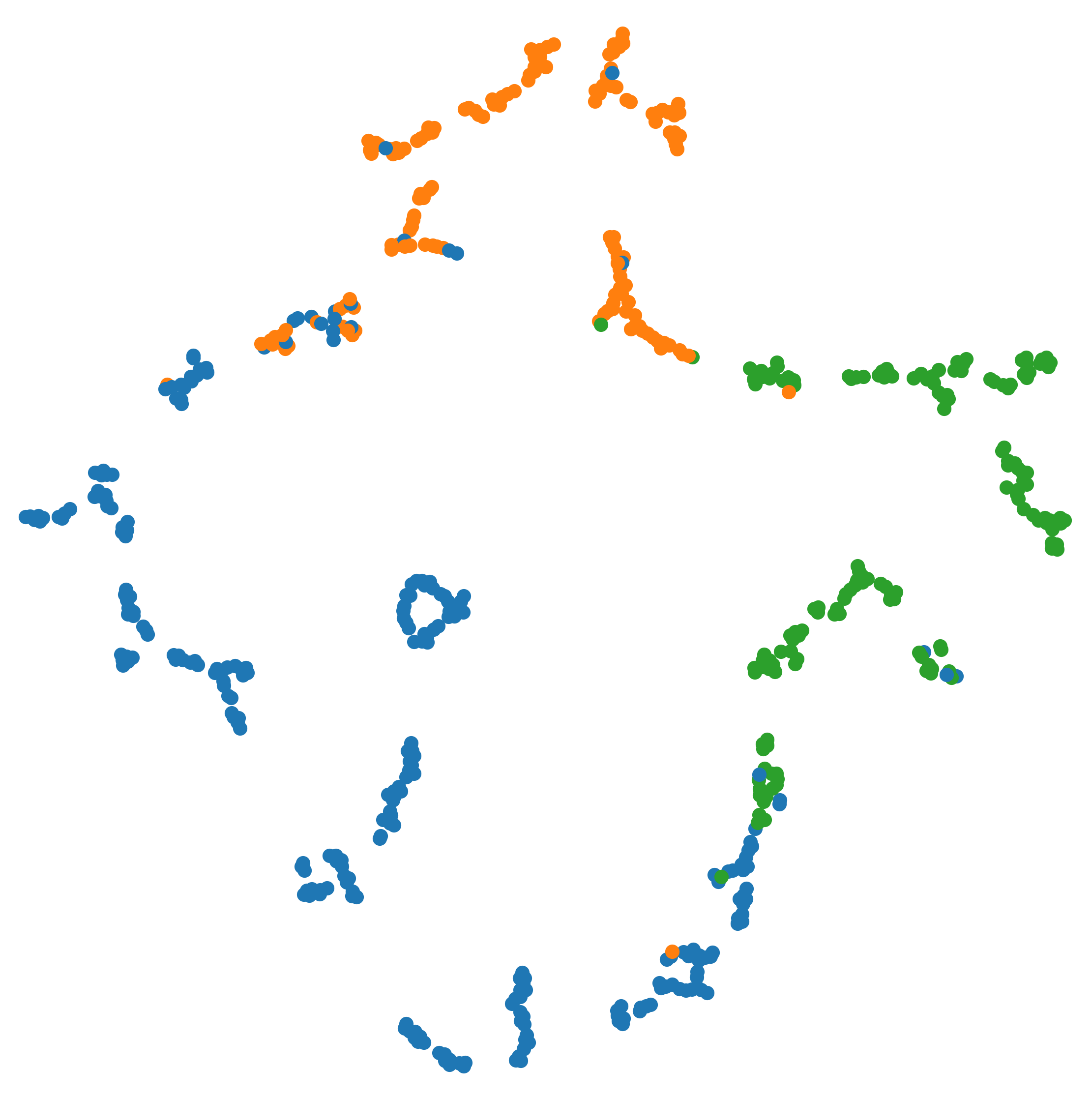}
  \end{minipage}
  \caption{Visualization of feature distribution within our proposed MediAD. (a) 3D Visualization on NACC dataset. (b) 2D Visualization on NACC dataset. (c) 3D Visualization on ADNI dataset. (d) 2D Visualization on ADNI dataset.}
  \label{fig:tsne}
\end{figure*}

To geometrically characterize the framework's class discriminability, we performed nonlinear dimensionality reduction using t-distributed stochastic neighbor embedding (t-SNE)~\cite{tsne}, projecting high-dimensional feature representations from both NACC and ADNI datasets into three-dimensional and two-dimensional latent spaces.
As shown in Fig.~\ref{fig:tsne}, we use different colors to denote the ground-truth labels of the data points, where blue points represent CN, orange points represent MCI, and green points represent AD. For the three-class classification task on the ADNI dataset, the data points from different classes exhibit good separability. On the NACC dataset, the separability between CN and MCI data points is poorer compared to their separation from AD points, which is consistent with our experimental results.

\section{Conclusions}
\label{Conclusions}
In this paper, we utilize the Structural Causal Model to model the causal relationship between clinical input and Alzheimer's Disease diagnosis. Based on this foundation, we propose MediAD, a causality-inspired framework for Alzheimer's diagnosis through multi-modal information. We implement causal intervention to mitigate the effects of the confounder that can lead to misdiagnosis. 
Specifically, inspired by the mediation analysis utilized in clinical reasoning, our MediAD introduces a cross-modal Causal Fusion (CF) module, a Front-Door Adjustment (FDA) module, and consistency loss ($\mathcal{L}_{cl}$) to enhance diagnostic reliability. 
Pre-trained on the NACC and ADNI datasets, our method demonstrates competitive performance compared to other methods.
Our future work will focus on investigating more sophisticated and theoretically sound designs for the mediator, as well as extending the application of this framework to other domains.


\end{document}